\documentclass[letterpaper]{article} 
\usepackage{aaai24}  
\usepackage{times}  
\usepackage{helvet}  
\usepackage{courier}  
\usepackage[hyphens]{url}  
\usepackage{graphicx} 
\usepackage{natbib}  
\usepackage{caption} 
\usepackage{algorithm}
\usepackage{algorithmic}
\usepackage{amsmath}
\usepackage{xcolor}

\usepackage{newfloat}
\usepackage{listings}
\DeclareCaptionStyle{ruled}{labelfont=normalfont,labelsep=colon,strut=off} 
\floatstyle{ruled}
\newfloat{listing}{tb}{lst}{}
\floatname{listing}{Listing}
\title{Discretionary Trees: Understanding Street-Level Bureaucracy via Machine Learning}
\author {
    Gaurab Pokharel\textsuperscript{\rm 1},
    Sanmay Das\textsuperscript{\rm 1},
    Patrick J. Fowler\textsuperscript{\rm 2}
}
\affiliations {
    \textsuperscript{\rm 1}George Mason University\\
    \textsuperscript{\rm 2}Washington University at St. Louis\\
    gpokhare@gmu.edu, sanmay@gmu.edu, pjfowler@wustl.edu
}

\usepackage{bibentry}

\begin{document}

\maketitle

\begin{abstract}
Street-level bureaucrats interact directly with people on behalf of government agencies to perform a wide range of functions, including, for example, administering social services and policing. A key feature of street-level bureaucracy is that the civil servants, while tasked with implementing agency policy, are also granted significant discretion in how they choose to apply that policy in individual cases. Using that discretion could be beneficial, as it allows for exceptions to policies based on human interactions and evaluations, but it could also allow biases and inequities to seep into important domains of societal resource allocation. In this paper, we use machine learning techniques to understand street-level bureaucrats' behavior. We leverage a rich dataset that combines demographic and other information on households with information on which homelessness interventions they were assigned during a period when assignments were not formulaic. We find that  caseworker decisions in this time are highly predictable overall, and some, but not all of this predictivity can be captured by simple decision rules. We theorize that the decisions not captured by the simple decision rules can be considered applications of caseworker discretion. These discretionary decisions are far from random in both the characteristics of such households and in terms of the outcomes of the decisions. Caseworkers typically only apply discretion to households that would be considered less vulnerable. When they do apply discretion to assign households to more intensive interventions, the marginal benefits to those households are significantly higher than would be expected if the households were chosen at random; there is no similar reduction in marginal benefit to households that are discretionarily allocated less intensive interventions, suggesting that caseworkers are using their knowledge and experience to improve outcomes for households experiencing homelessness.

\end{abstract}

\section{Introduction}

Homelessness is an acute and pervasive issue affecting societies worldwide. The lack of stable housing and access to resources poses immense challenges for individuals and families, jeopardizing their well-being and livelihoods. 
As federal guidelines increasingly emphasize ending homelessness entirely \cite{all_in_2022}, a need exists for a comprehensive understanding of its causes, dynamics, and the strategic allocation of limited resources to provide timely interventions. In the United States, decisions on homeless service allocation are determined through a complex mix of federal guidelines, community and local agency policies, and the interpretation and implementation of these policies by on-the-ground caseworkers. 

Homelessness caseworkers are an example of what Lipsky, in a seminal work, calls {\em street-level bureaucrats}~\citep{Lipsky_2010}. These are public-facing employees of government agencies who are granted discretion and may (or may not) follow the agencies' policies in each decision they make. Among the reasons Lipsky details for why street-level bureaucrats implement different policies include resource limitations (e.g., the time and information constraints faced by tremendously overworked caseworkers and street-level bureaucrats), as well as the misalignment of individual and agency objectives that may be ambiguous.

This paper aims to use machine learning tools to better understand street-level bureaucrats' behavior. In particular, we are interested in three main questions grounded in the street-level bureaucracy theory: (1) Can bureaucratic decision-making be boiled down to applying a relatively simple set of heuristics/rules? The theory says street-level bureaucrats often adopt such methods to reduce cognitive load and decision-making time. (2) When decisions are not driven by these rules (examples of the application of \emph{caseworker discretion}), are caseworkers nonetheless \emph{consistent} in their decision-making? Inconsistent application of discretionary powers is a major problem because it leads to individuals suffering allocative harms without a procedural basis and could cause considerable bias and inequity. (3) When caseworker discretion is applied, what can we say about the cost/benefit trade-off to the subjects of such discretion?

We leverage a rich dataset of allocations of households to homelessness services during a period in which the allocation process was less formulaic than imposed by the subsequent adoption of decision-making tools \cite{eubanks2018automating, hud_2017_coc}. In addition to the interventions assigned, the dataset contains demographic and other features of the households. We use machine learning methods to ask the question \emph{how predictable are the interventions assigned to households} and to understand the nature of caseworker decisions. We present evidence to support several important claims related to the questions above. First, simple rules can capture some, but not all, of the decision-making about which allocations households receive. State-of-the-art ``short'' decision trees achieve areas under the ROC curve (AUCs) of $0.76$ and $0.87$ when predicting whether households will receive Emergency Shelter and Transitional Housing, respectively (the two main interventions we study). Second, caseworker decision-making is generally highly consistent beyond what can be explained with simple decision trees. The AUCs for gradient-boosted trees on the same two tasks are $0.95$ and $0.94$, respectively. Thus, caseworkers are procedurally consistent in their decision-making. Third, caseworkers apply discretion  (operationalized as decisions that short decision trees would not have correctly predicted) in an interesting way. Keeping in mind that Transitional Housing (TH) is a more intensive intervention that generally prevents reentry into homeless services (the outcome of interest) better than receiving less intensive Emergency Shelter (ES) that has poorer outcomes, we find that caseworkers typically only apply discretion (in either direction: moving households predicted to do best in TH to ES or predicted ES to TH) on households that would be considered less vulnerable. However, the group that moves from predicted ES to TH (thus, to the more intensive intervention) systematically receives higher expected marginal benefit from this change; conversely, the group that moves from TH to ES does not receive significantly lower expected marginal benefit. These estimates of marginal benefit are based on counterfactual machine learning predictions and were not available in any form to the caseworkers; thus, this third result implies caseworkers can assess and apply discretion in a highly sophisticated manner to the overall benefit of vulnerable populations.

More broadly, the work represents the first to our knowledge that applies techniques from artificial intelligence and machine learning to understand decision-making in the context of street-level bureaucracies, which are specially tasked with many of the core public-facing tasks of local and state governments. Shedding light on how human decision-making works may help us understand how and when automated decision-support may be most valuable to those charged with interfacing with the public in high-stakes allocation and justice scenarios. By gaining a deeper understanding of how organizational policies interact with the street-level bureaucracy typically employed by caseworkers, we can foster greater clarity and transparency in resource distribution processes and start to untangle some of the intricate dynamics between organizational policies and the caseworkers responsible for implementing them.

\section{Background and Related Work}

\paragraph{Homelessness}
Federal guidelines define homelessness as residence in unstable and non-permanent accommodations. This includes emergency shelters, temporary housing, and places not meant for habitation, such as cars, parks, and abandoned buildings (P.L. 112-141). According to \citet{HUD22_report}, more than 580,000 people experienced homelessness on a single night in January 2022 across the United States, while approximately 1.5 million people use homeless services at some point during each year \cite{HUD18_report}. Furthermore, three in ten people experiencing homelessness did so as a part of a family with children under 18 years of age, and one-third of individuals experiencing homelessness exhibited chronic patterns of homelessness \cite{HUD22_report}. The enduring consequences of homelessness and the resulting upheaval have lifelong implications with substantial societal burdens in terms of reduced productivity, compromised health, and increased expenditures on compensatory social services \cite{khadduri2010costs,culhane2011patterns,fowler_complex_systems}.

There has been some recent work on AI approaches towards homelessness. Much of this work has focused on techniques designed to improve resource allocation to homeless households \cite{azizi2018designing, JAIR, Kube_Das_Fowler_2019}. Other related work has examined what values and efficiency/fairness tradeoffs have been \cite{FAccT} or should be \cite{vayanos2020robust} encoded into algorithmic approaches. Despite the pivotal role played by allocation decisions in shaping outcomes for homeless individuals and families, substantial gaps remain in understanding the deployment of homeless services \cite{VISPDAT_validity,capability_traps,efficient_targeting_shinn}. The knowledge gap hinders the development of evidence-based strategies and the establishment of equitable mechanisms for resource allocation within homeless services. 
As far as we know, this is the first paper to use machine learning approaches to examine caseworker decision-making.



\paragraph{Street-level bureaucracy}
The seminal work of \citet{Lipsky_2010} on street-level bureaucracy is the theoretical underpinning for the questions we ask in this paper. As mentioned in the Introduction, Lipsky notes the tensions between different caseworker objectives and the importance of their ability to use judgment and discretion in their sense of self. It is precisely these tradeoffs we explore in this paper.

Some recent qualitative work has used the lens of street-level bureaucracy in thinking about human-AI interaction. \citet{childWelfare2022} conduct a detailed qualitative investigation into how child welfare workers utilize an AI Decision Support (ADS) system called the Allegheny Family Screening Tool (AFST) in their day-to-day decision-making processes. The study finds that workers often perceive value misalignments between the ADS's predictive targets and their own objectives. Workers compensated for the limitations of the ADS by relying on their contextual knowledge of individual cases. Organizational pressures and incentives influenced workers' reliance on the ADS, sometimes leading them to comply with the AFST recommendations even against their best judgment. Additionally, workers felt limited agency in shaping the ADS's use or improving its accuracy. The paper highlights the need to address human-AI value misalignments.

\citet{homelessnesPerspective2023} conducted an in-depth study using AI lifecycle comicboarding, a novel method that allowed frontline workers and unhoused individuals to provide specific feedback on an AI system's design and deployment. Most pertinently to our work, participants expressed concerns about potential biases and harm caused by the algorithm, particularly in its optimization towards the county's interests rather than addressing the needs and safety of unhoused individuals. Our work complements this literature thread by analyzing human decision-making (rather than human-AI complementarity) using tools from AI (rather than qualitative methods). It thus sheds a ``revealed preference'' style light on what caseworkers do in practice.

\paragraph{Interpretability}

Our main connection to the interpretability literature is how we operationalize our notion of identifying short heuristic rules that caseworkers could use to make quick decisions, reducing cognitive load in ``easy'' cases. We do so using a relatively new short decision-tree learning algorithm, which comes from interpretable machine learning. \citet{Doshi_Kim_2017} highlights the importance of interpretable machine learning and advocates for a scientific approach. The field of interpretability often focuses on efforts to explain model characteristics, like the LIME framework \cite{LIME}, SHAP values \cite{SHAP}, and risk scores \cite{rudin_2019}. But using such black-box, model-independent metrics -- while offering valuable insights into opaque models -- suffers from limitations in the specific context of what we are trying to accomplish. The main issue is that their model-agnostic nature may lead to explanations that fail to capture the idiosyncrasies of specific models, potentially compromising the accuracy of the interpretations. In this study, we embrace Decision Trees (DT) as the model of choice due to their inherent interpretable nature, the importance of which has been stressed by \citet{rudin_2019} when making high-stakes decisions. DTs' model-specificity ensures accurate representations of individual models, providing insights into feature contributions and decision rules. We use a recent algorithmic technique proposed by \citet{SERDT} to learn short decision trees to try and enhance our understanding of caseworker behavior.

\subsection{The Data}

Through a data sharing agreement, this work uses administrative records from the homeless management information system (HMIS) in St. Louis, MO, collected from 2007 through 2014. Local service providers collect data in real-time using a web-based platform as individuals and families seek federally funded homelessness assistance. The platform is managed by a local non-profit organization contracted with the homeless system, which offers support, user training, technical assistance, and quality control. 
 
The HMIS maintains detailed records, encompassing information on household-level characteristics, such as demographics, housing risk, and eligibility determinations. Additionally, the data captures entry and exit dates from five federally defined types of homeless assistance in increasing order of their intensity: homelessness prevention (prev), emergency shelter (ES), rapid rehousing (RRH), transitional housing (TH), and permanent supportive housing. The metropolitan area operated a telephone hotline to coordinate service delivery. Records exist on every call, including dates and service referrals. Through the use of household identifiers, information can be linked across different time periods. Data-sharing agreements with regional homeless systems are in place to ensure privacy and security, allowing access to deidentified records in accordance with Institutional Review Board guidelines. Best practices in data security, including ethics training, are followed for data transfer, storage, and analysis.

\subsection{Data Cleaning}

Service records come from 75 different housing agencies serving households assigned a unique, anonymous household identification number \citet{JAIR}. Data contains household characteristics available upon entry into the system, as well as information on all entries and exits from different homeless services. We exclude permanent supportive housing for the present study because the service was rarely used as an initial response for first-time entries into the homeless system between 2007 and 2014. In contrast to \citet{JAIR} that assessed whether households reentered the homeless system within a two-year follow-up, we aim to predict the interventions assigned to the households. Hence, our target variables are one of the four remaining intervention types, namely: TH, RRH, ES, and Prev. We use a total of 34 features to predict these targets, some of which are listed in Table \ref{table:feature_summary} with a complete list provided in the supplementary material.

\begin{table}[H]
\resizebox{\columnwidth}{!}{%
\begin{tabular}{|c|c|c|}
\hline
Feature Type                    & Number of Features & Examples                                             \\ \hline
Binary Features                 & 3                 & Gender, Spouse Present, HUD Chronic Homeless         \\
Non-Binary Categorical Features & 17                 & Veteran Status, Disabling Condition, Substance Abuse \\
Continuous Features             & 14                 & Age, Monthly Income, Wait Time, Calls to Hot-line     \\ \hline
Total Features                  & 34                &                                                      \\ \hline
\end{tabular}%
}
\caption{Summary of features in the dataset}
\label{table:feature_summary}
\end{table}
 
\noindent As Table \ref{table:feature_summary} illustrates, we used a collection of binary, categorical, and continuous features. We one-hot-encoded the categorical features. We refer the reader to the supplementary material for details regarding the characteristics of the data, including the makeup and distribution of the interventions assigned.

\paragraph{Time context:} As mentioned, the data captures first-time entries before 2014. At this time, there was no explicit formula for prioritization for different interventions used by the homeless service system. Thus, caseworkers relied more on their own judgment in making decisions. They balanced potential trade-offs between prioritizing based on household vulnerability (e.g., chronically homeless households with comorbid conditions) and outcome -- or allocations expected to be the most beneficial for households. Federal policy shifts following the study period required the adoption of coordinated entry into services based on vulnerability scores, such as measured by the Vulnerability Index-Service Prioritization Decision Assistance Tool (VI-SPDAT) \cite{orgcode2015vulnerability}.

\paragraph{Intervention focus} While we report results on the accuracy of simple decision rules and the consistency for all four interventions we study, we focus on two groups -- Transitional Housing (TH) and Emergency Shelter (EH) -- in analyses. The other two interventions (RRH and Prev) were only available for a part of the study period, as they were funded through the American Recovery and Reinvestment Act in the wake of the 2008 Great Recession. Other work has shown that Prevention, in particular, is a different type of intervention that is not directly comparable to the others in terms of eligibility and outcomes \cite{JAIR}. 

\begin{figure}[H]
    \centering
    \includegraphics[width=\columnwidth]{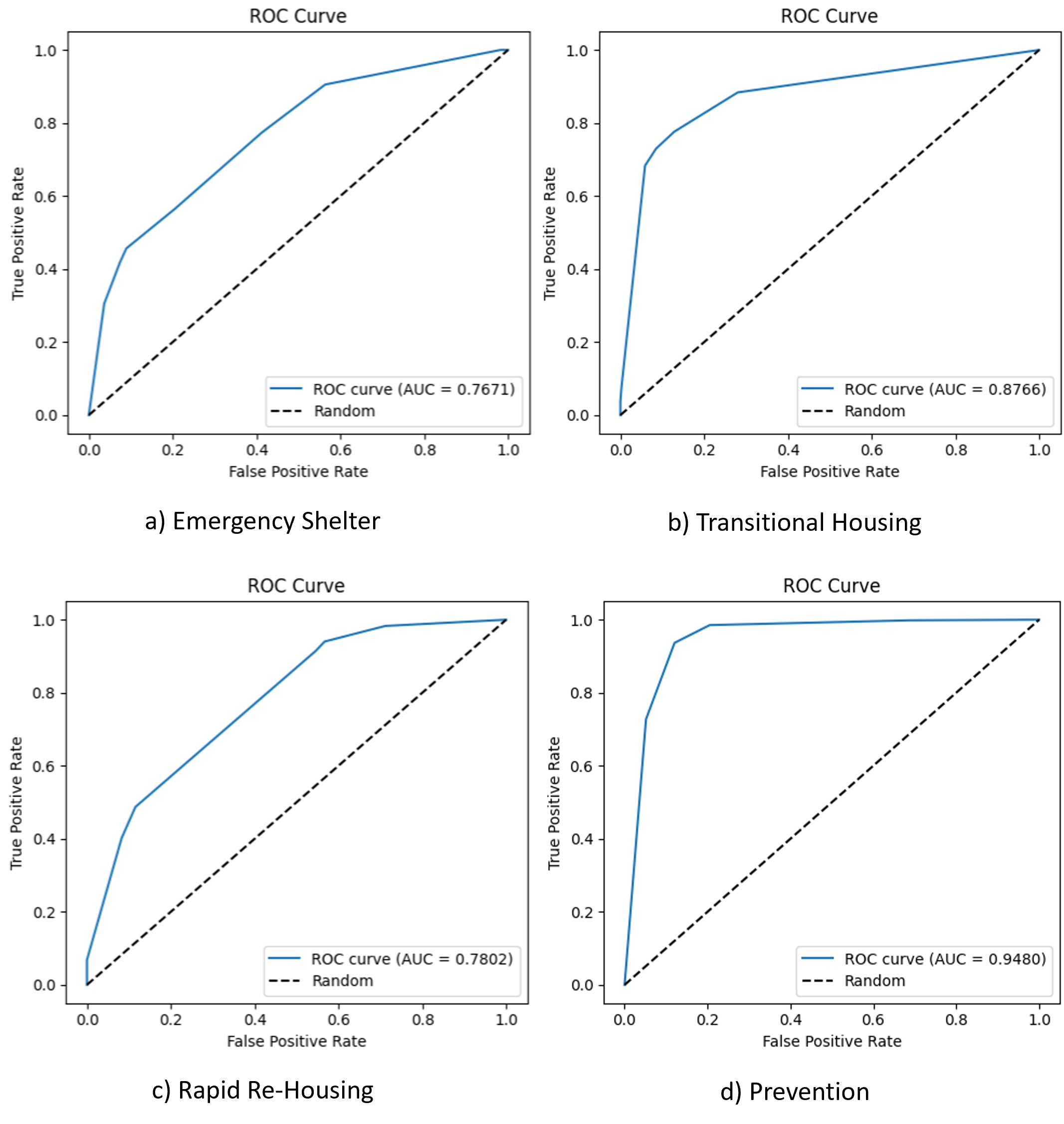}
    \caption{ROC curves for prediction of intervention assignment in a one-vs-all setting for all four interventions using SER-DT. Simple rules are quite effective at predicting interventions overall, implying the use of short rules/heuristics by caseworkers in many cases is plausible and consistent, as predicted by theory.}
    \label{fig:roc_curves_short_DTs} 
\end{figure}

\begin{figure*}[ht]
    \centering
    \includegraphics[width=\textwidth]{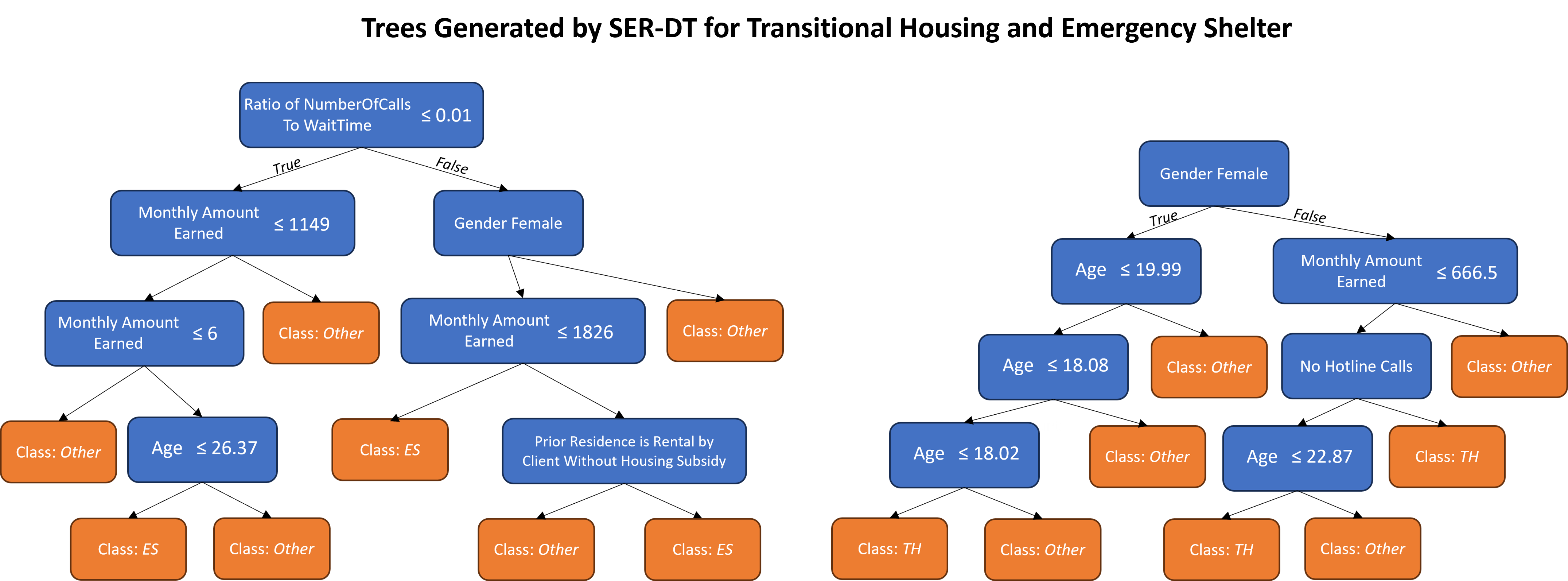}
    \caption{The decision trees generated by SER-DT for the assignment of Emergency Shelter (left) and Transitional Housing (right). The left branches of each node indicate that the parent condition was satisfied, whereas the right branch indicates the condition was false. The generated trees are small but can encompass a significant portion of the correctly made decisions.}
    \label{fig:serdt_trees} 
\end{figure*}
\section{Results: Simplicity and Consistency in Intervention Assignment}
Our main investigatory tool is the predictability of caseworker behavior in assigning households to particular interventions. We employ ML methods to learn the associated decision rules. We start by investigating what fraction of caseworker behavior can be captured by simple decision rules before considering the consistency in intervention assignments.

\subsection{Simple Decision Rules}
Street-level bureaucracy theory suggests that caseworkers develop simple heuristics that enable them to quickly make decisions on most cases, reserving their expertise and time for more complex cases. To test this, we attempt to learn intervention assignments using \emph{short trees}. These are functionally equivalent to simple rules, or the common understanding of heuristics in the behavioral sciences, which are typically contrasted with more cognitively demanding linear models \cite{Hogarth_Karelaia_2007}. We use a relatively new algorithm from the literature, the SER-DT algorithm \cite{SERDT}. This algorithm is specifically designed to optimize ``explanation size'' -- a custom metric aimed at enhancing explainability using as simple a decision tree as possible.

We applied the method to one-vs-all prediction for the four interventions described above. A partition ratio of 70\% for training data and 30\% for testing data was employed. Analogous to established methodologies, such as the Classification and Regression Trees (CART) algorithm, the SER-DT method identifies splits that yield sub-trees characterized by diminished `impurity’. Unlike other algorithms, SER-DT's inherent non-greedy nature causes minor variations across runs. This occurs because SER-DT weights the Gini impurity at each node using a hyper-parameter called `FactorExpl' (which was set to 0.97, mirroring the practice from the original paper for a minimum trade-off in accuracy when seeking explainability), and thus, impacting which split gets used on which feature. To ameliorate this slight variation in trees, 10 trees were trained with distinct seeds per intervention. We report the average AUC values of the trees for each intervention in Table \ref{table:sertdt_auc} along with the AUC values for the best model (taken as the model with the highest test accuracy) and its 95\% confidence interval (CI) calculated using DeLong's method \cite{DeLong_DeLong_Clarke-Pearson_1988} with a sample size of 1907. Moreover, the structural depth of the trees is constricted to a maximum limit of four tiers. This constraint is formulated in congruence with the guiding principle of preserving the ``simplicity’’ of the decision rules. Greater depth gains higher accuracy at the expense of potentially making the tree more complicated to understand.

\begin{table}[]
\resizebox{\columnwidth}{!}{%
\begin{tabular}{|c|c|c|c|}
\hline
\textbf{Intervention} & \multicolumn{1}{l|}{\textbf{Average AUC Value}} & \textbf{AUC (Best Model)} & \textbf{95\% CI (Best Model)} \\ \hline
\textbf{Emergency Shelter}    & 0.7608 & 0.7671 & {[}0.7452, 0.7890{]} \\
\textbf{Transitional Housing} & 0.8703 & 0.8766 & {[}0.8547, 0.8984{]} \\
\textbf{Rapid Re-Housing}     & 0.6508 & 0.7802 & {[}0.7463, 0.8170{]} \\
\textbf{Prevention}           & 0.9384 & 0.9480 & {[}0.9371, 0.9589{]} \\ \hline
\end{tabular}%
}
\caption{Average AUC values over 10 different runs, AUC values for the best model, and the 95\% confidence interval for the best model predicting different interventions using SER-DT.}
\label{table:sertdt_auc}
\end{table}


Figure \ref{fig:roc_curves_short_DTs} shows the AUC curves for the best performing model in each case, while the corresponding trees generated specifically for TH and ES appear in Figure \ref{fig:serdt_trees}. As can be seen, the short rules are good at explaining a chunk of human decision-making regarding intervention assignments based on household characteristics. We see that the short decision tree achieves an AUC of $0.87$ in predicting whether a household will receive TH. Examining the tree suggests the basic rules for achieving the high AUC: Assign TH to households with male heads who make \$666.5 or less and either have prior calls to the hotline or are young (less than 23yo); assign TH to households with young (less than 19 years) female heads of household. The assignment of ES is somewhat less predictable by simple rules (AUC of $0.76$). The rules, again, are interesting when operationalizing the ratio of the number of calls to wait time as a notion of ``urgency'' for services. One fundamental rule is that households with female heads who have at least some urgency are sent to ES unless their monthly earnings are relatively high (greater than \$1826) \textbf{and} their immediate prior residence was an unsubsidized rental. Together with the TH rules, this implies that males are generally prioritized in the simple heuristic rules for transitional housing, possibly because of comorbidities (like alcohol or drug abuse, disability, mental health issues -- see \citet{JAMIA} for a discussion).

\begin{figure}
    \centering
    \includegraphics[width=\columnwidth]{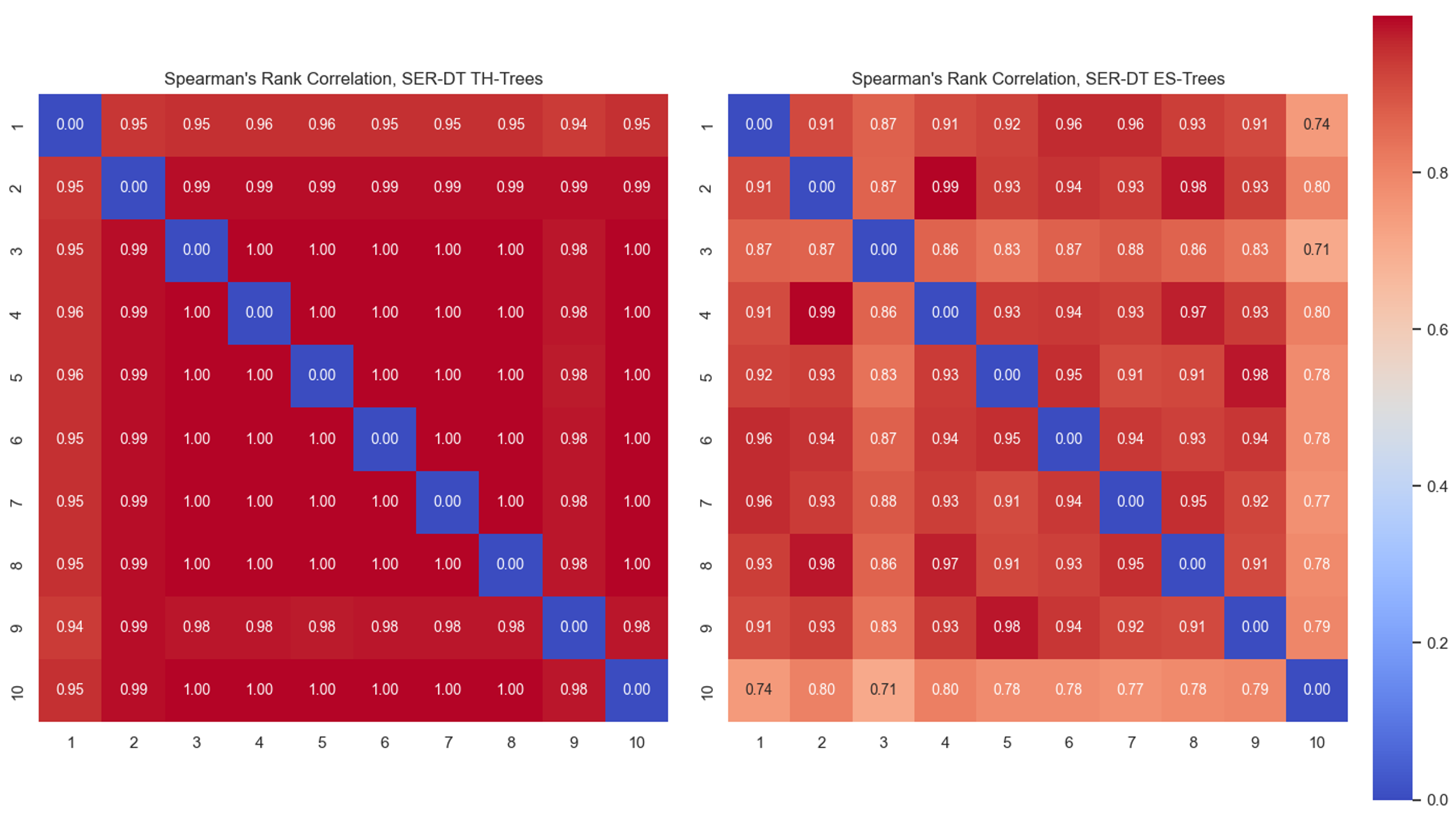}
    \caption{Spearman's rank correlation for ten different models predicting Transitional Housing (left) and Emergency Shelter (right). We observe a high degree of correlation in model output, alleviating any concern regarding differences in tree outcomes.}
    \label{fig:correlations}
\end{figure}

For our analysis in the following sections, we take our predicted intervention by SER-DT to be the labels generated by the best performing models i.e. the ones with the highest test accuracy. 
This could raise a concern about the identifiability of human decisions -- for example, how different are predictions made by the same model (SER-DT) due to training set variation? It is important to note that the key question here is about the implications of any model differences on the population they predict would receive different interventions, rather than about differences in models themselves. To test this, we compute rank correlations \cite{Spearman_1904} on ten different SER-DT models, demonstrating a high degree of similarity in the outputs of different models for both TH and ES, affirming consistency in model predictions and alleviating concerns about differences in tree outcomes (see Figure \ref{fig:correlations}).


\subsection{Complex Decision Making}

To understand whether caseworkers consistently assign interventions even when not using simple decision rules, we investigate the predictivity achievable when using a more complex machine learning algorithm. The algorithm of choice here was XGBoost -- an ensemble learning algorithm that has gained prominence for its exceptional predictive performance and versatility in a wide range of ML tasks.\footnote{We also do a similar analysis using the CART algorithm. CART has comparable performance to XGBoost in terms of productivity. See supplementary material for further details.} Again, the dataset was partitioned into training and testing sets, with a ratio of 70\% for training and 30\% for testing. A systematic grid search approach was employed to optimize hyper-parameters. The explored parameters included the number of estimators (ranging from 50 to 20, increasing in steps of 50) and the maximum depth of the trees (ranging from 2 to 8, increasing in steps of 2). 

\begin{figure}
    \centering
    \includegraphics[width=\columnwidth]{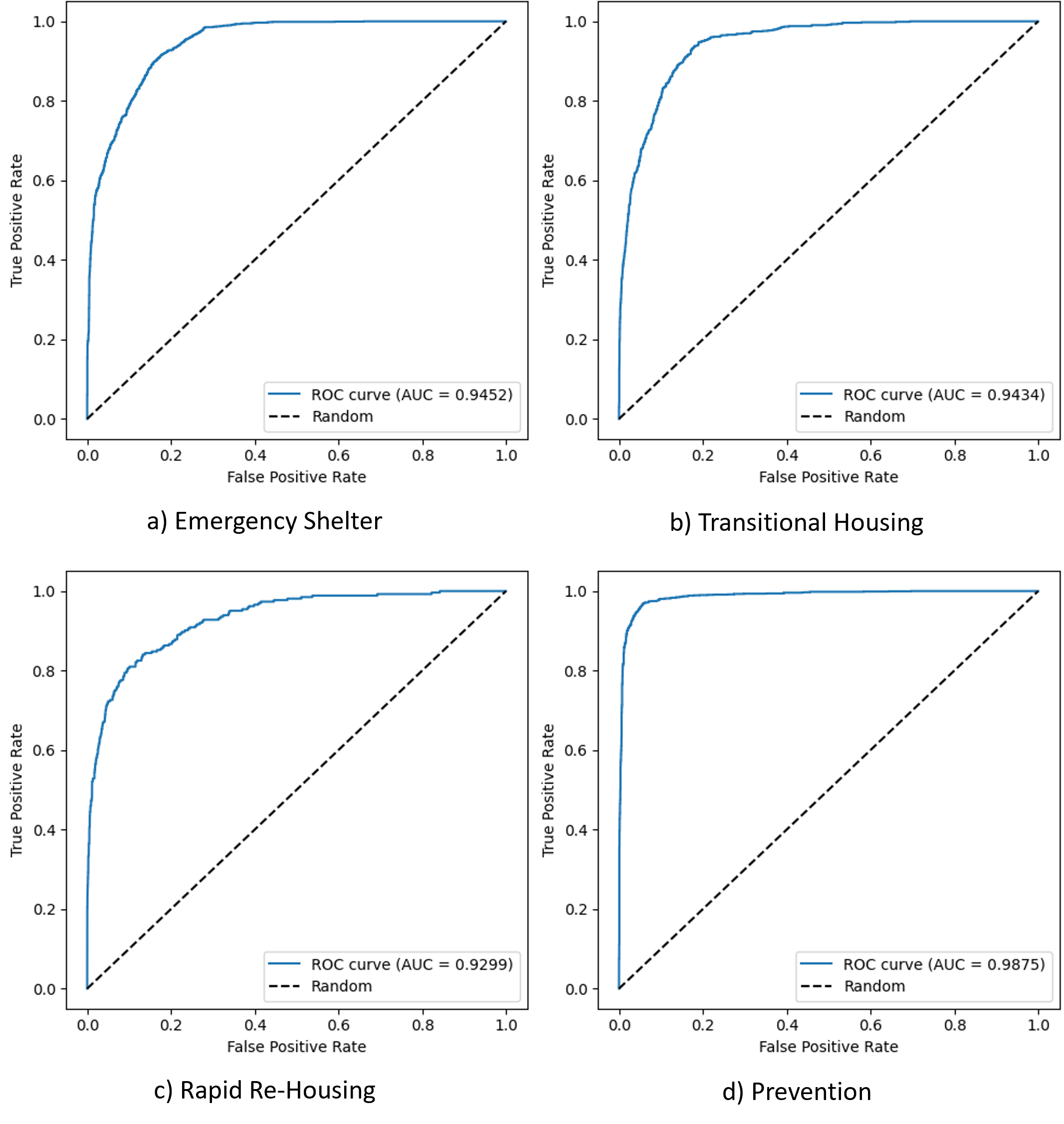}
    \caption{ROC curves for each tree used to learn intervention assignments using the XGBoost algorithm. Predictive discrimination, measured using the area under the ROC curve is very high, implying significant consistency in caseworker decision-making.}
    \label{fig:roc_curves_individual_DTs} 
\end{figure}

The optimal hyper-parameters varied across interventions. Specifically, for TH, the optimal configuration was a max depth of 2 with the number of estimators equaling 150. For ES, a max depth of 4 with the number of estimators set to 100 yielded the best performance. RRH achieved optimal results with a max depth of 6 and with the number of estimators at 50, while Prev benefited from a max depth of 2 and the number of estimators set to 200. Figure \ref{fig:roc_curves_individual_DTs} displays the ROC curves for the individual decision trees and Table \ref{table:xgb_auc} provides details on AUC values and confidence intervals (with a sample size of 4239).

\begin{table}[]
\begin{tabular}{|c|c|c|}
\hline
\textbf{Intervention}         & \textbf{AUC value} & \textbf{95\% AUC CI} \\ \hline
\textbf{Emergency Shelter}    & 0.9452             & {[}0.9392, 0.9513{]} \\
\textbf{Transitional Housing} & 0.9434             & {[}0.9360, 0.9507{]} \\
\textbf{Rapid Re-Housing}     & 0.9299             & {[}0.9136, 0.9462{]} \\
\textbf{Prevention}           & 0.9875             & {[}0.9846, 0.9904{]} \\ \hline
\end{tabular}
\caption{AUC values and their 95\% confidence intervals for the trees used to learn intervention assignments using the XGBoost algorithm.}
\label{table:xgb_auc}
\end{table}

Figure \ref{fig:roc_curves_individual_DTs} illustrates that the decision rules implemented by caseworkers can be effectively learned with high predictive accuracy. The findings demonstrate that caseworkers are \emph{consistent} in their decision-making. This is important for concerns about procedural justice since there is little arbitrariness in the application of human judgment in these cases. This then raises the question of why caseworkers choose to apply their discretion in these cases, and what it likely means for the outcomes of intervention assignment.

\section{Results: Discretionary Assignments}

What explains the application of discretion in the significant fraction of cases where the assignment mismatches that which would have been predicted by the simple decision rule? In order to better understand this, we consider and operationalize two possibilities that have been hypothesized as major factors in human decision-making on the allocation of scarce homelessness resources: \emph{vulnerability} and \emph{outcome} \cite{localjustics_Das_2022}.  

\subsection{Measuring Vulnerability and Potential Outcomes}


To begin, let's delve into the concept of vulnerability. The Vulnerability Index-Service Prioritization Decision Assistance Tool (VI-SPDAT) is as a widely employed metric to assess vulnerability within homelessness resource allocation \cite{orgcode2015vulnerability}. Despite its subsequent recognition for introducing biases against specific sub-groups \cite{VISPDAT_validity, cronley2022invisible, shinn2022allocating}, and its decreasing prevalence, it remained the conventional yardstick for prioritization during the period following our study. Given this context, if vulnerability indeed played a pivotal role in the application of discretion, we would anticipate its correlation with decision-making. While our dataset lacks direct VI-SPDAT scores, we endeavored to closely replicate it using available data to establish a vulnerability score (VS). The full point assignment system we developed is detailed in supplementary material.

Next, we turn to outcomes. While each household has a range of different possible outcomes, local agencies are often evaluated on the basis of success at helping families out of homelessness, that is, whether households were stably housed some time (typically two years) after the homelessness intervention. Therefore we directly use the probability of return to homelessness conditional on different interventions using Bayesian Additive Regression Trees, as estimated in prior work \cite{JAIR, JAMIA}. Note that the counterfactual estimates provide a sense of how much better one intervention is than another in terms of reducing the probability of returning to homelessness.

We note here again that there is an inherent tension between vulnerability and outcome. Both of these are classic metrics for prioritization in the local justice literature \cite{localjustics_Das_2022}, but it is recognized that they may often produce opposing motivations. Caseworkers frequently are asked to prioritize vulnerability by federal guidelines, while at the same time, the agency is evaluated and funded based on success in getting households out of homelessness \cite{fowler_complex_systems}. A very vulnerable household may not benefit much from an intensive intervention in some cases, but it may still be appropriate to prioritize them according to our values. These types of trade-offs are what caseworkers have to deal with on a daily basis \cite{kube2022just}.

\subsection{Investigating Discretion}

\begin{figure}[]
    \centering
    \includegraphics[width=\columnwidth]{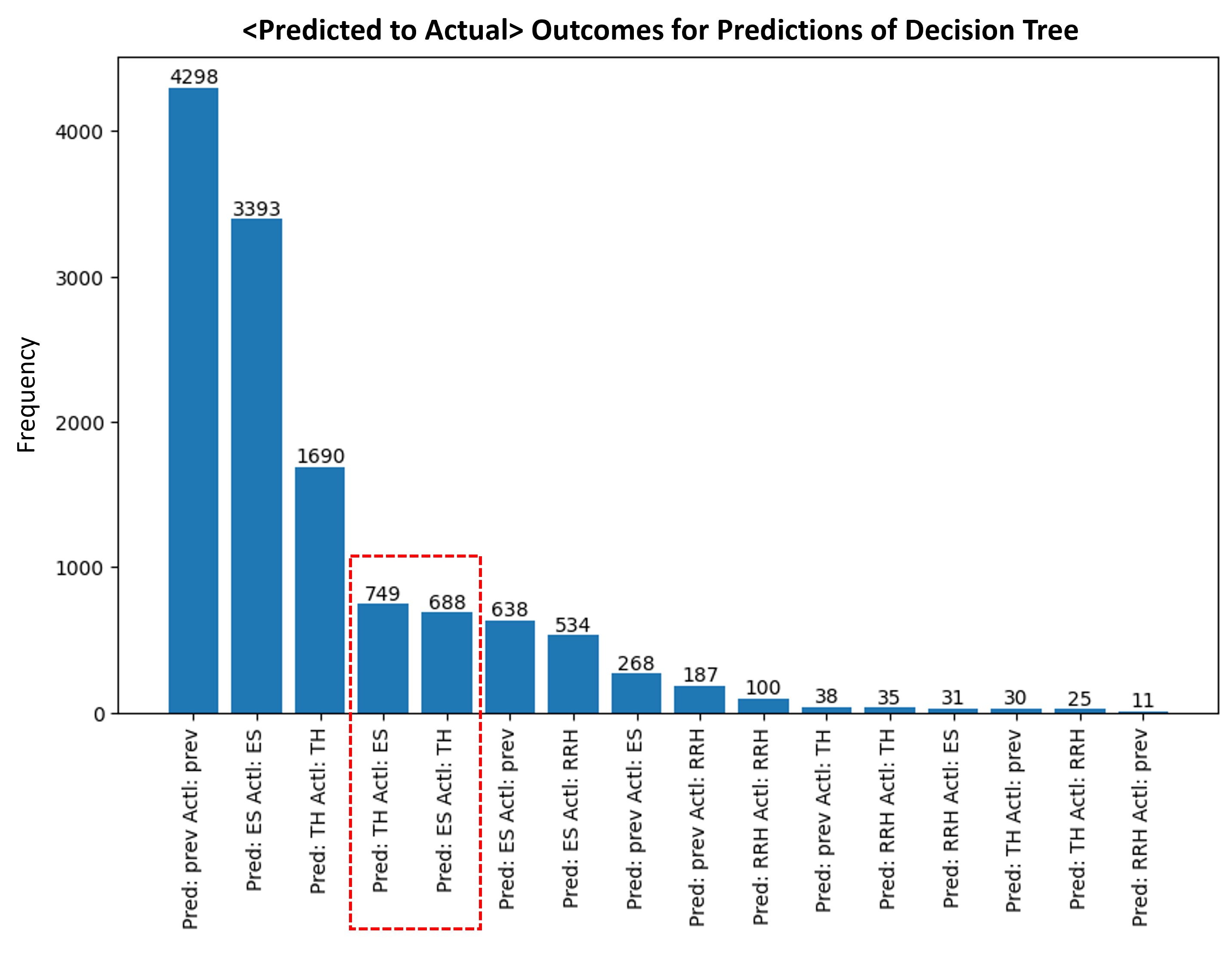}
    \caption{Distribution of the number of records for predicted versus actual assignment of intervention. We focus our analysis of discretion quantification in the subgroups marked within the red  box.}
    \label{fig:prediction_difference}
\end{figure}

Can we understand applications of discretion through the lens of vulnerability and/or outcome? We conducted a targeted analysis focusing on a subset of the data where the actual assignments differed from the SER-DT model predictions. Given the employment of a one-versus-all model for each intervention, there exists the potential for an individual household to be predicted to receive multiple interventions. In such instances, the final forecasted intervention for the household is taken from the prediction deemed most ``confident." Among the 12,546 data points, 9,481 instances aligned with the documented assignments in our dataset, while 3,065 data points (nearly one-quarter) showed disparities. Within the subset of mismatches, we further investigated two distinct groups: 749 data points where our model predicted TH, but ES was assigned; and 688 data points where ES was predicted, but TH was assigned. These subsets represent half of the data points within the discretionary subgroup and are characterized by a clear hierarchy of the intervention's intensity, with transitional housing being more intensive than emergency shelters. Analyzing the specific subgroups allows us to assess the presence and extent of caseworker discretion in intervention assignments. The full distribution of the predicted versus actual intervention assignments is displayed in Figure \ref{fig:prediction_difference}.

For simplicity, we call the group predicted to receive ES but actually receiving TH ``EStoTH'' and the group predicted to receive TH but actually receiving ES ``THtoES''. We compute vulnerability scores (VS) as described above for each household in the entire population. Additionally, we estimate the marginal benefit of being assigned TH rather than ES as $\text{MB} = \Pr (\text{return to homelessness} \mid \text{ES}) -  \Pr (\text{return to homelessness} \mid \text{TH})$. A nonparametric test was used to identify differences between the group actually receiving the discretionary assignment and a random group that would have received the assignment. For example, when considering VS, the test compares the average VS score of the EStoTH group with the distribution of average VS scores of 1000 different groups of the same size (688) sampled at random from the population that was predicted to receive ES in the first place. Similarly, the test compares the average MB score of, say, the THtoES group with the distribution of average MB scores of 1000 different groups of the same size (749) sampled at random from the population that was predicted to receive TH in the first place.

\subsubsection{Vulnerability}
\begin{figure}[]
    \centering
    \includegraphics[width=\columnwidth]{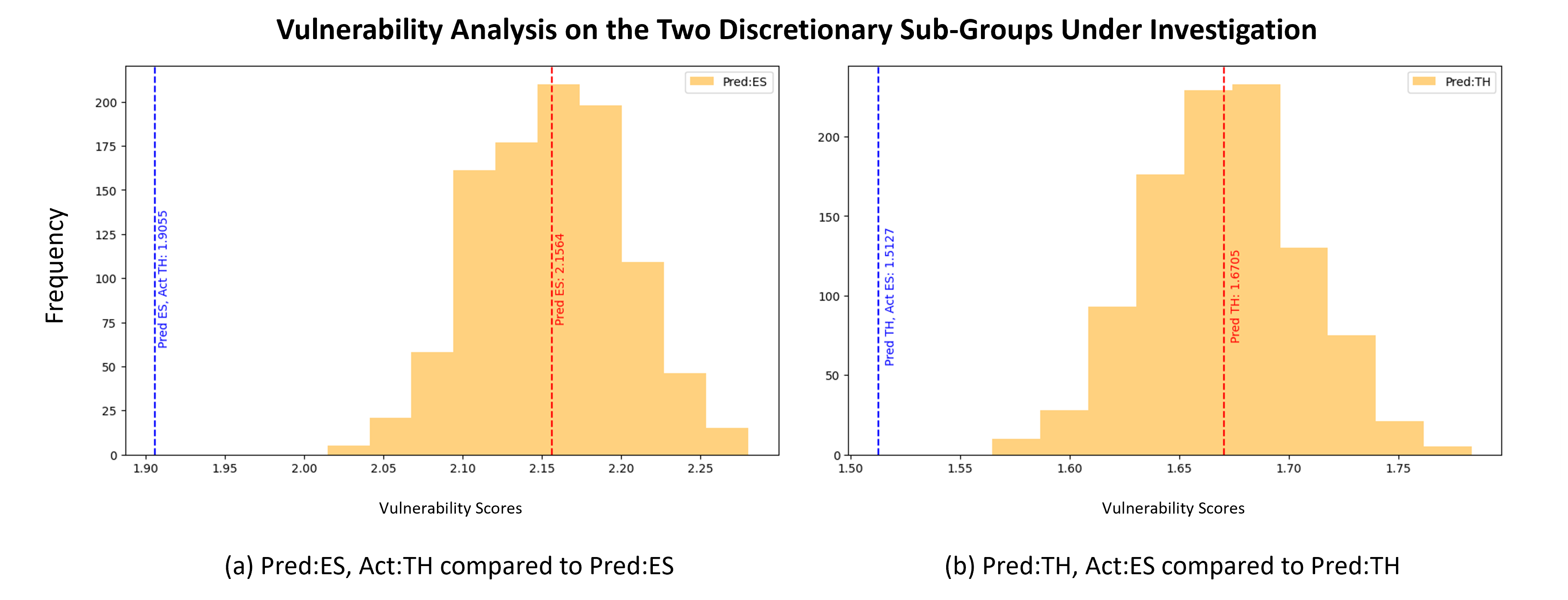}
    \caption{Non-parametric test comparing vulnerability scores between the discretionary subgroups and the population predicted to receive the intervention.}
    \label{fig:vulnerability_analysis}
\end{figure}

Figure \ref{fig:vulnerability_analysis} shows the results of the nonparametric tests for VS. For both the group predicted to receive TH but ultimately assigned ES, and the group predicted to receive ES but ultimately assigned TH, the vulnerability scores of the discretionary subgroups were significantly lower than one would expect if they were randomly sampled from the population of those predicted to receive the (respective) original intervention. The graphs themselves show a histogram (in orange) of the distribution of the 1000 means obtained from the control group. Within the histogram, the vertically dotted line in red represents the ``mean of the means'' for the control group, while the vertically dotted blue line represents the mean for the actual discretionary group, lying well outside the sampled distribution in both cases. This demonstrates quite clearly that caseworkers are only applying discretion in the context of especially low-vulnerability households.

\begin{figure}[]
    \centering
\includegraphics[width=\columnwidth]{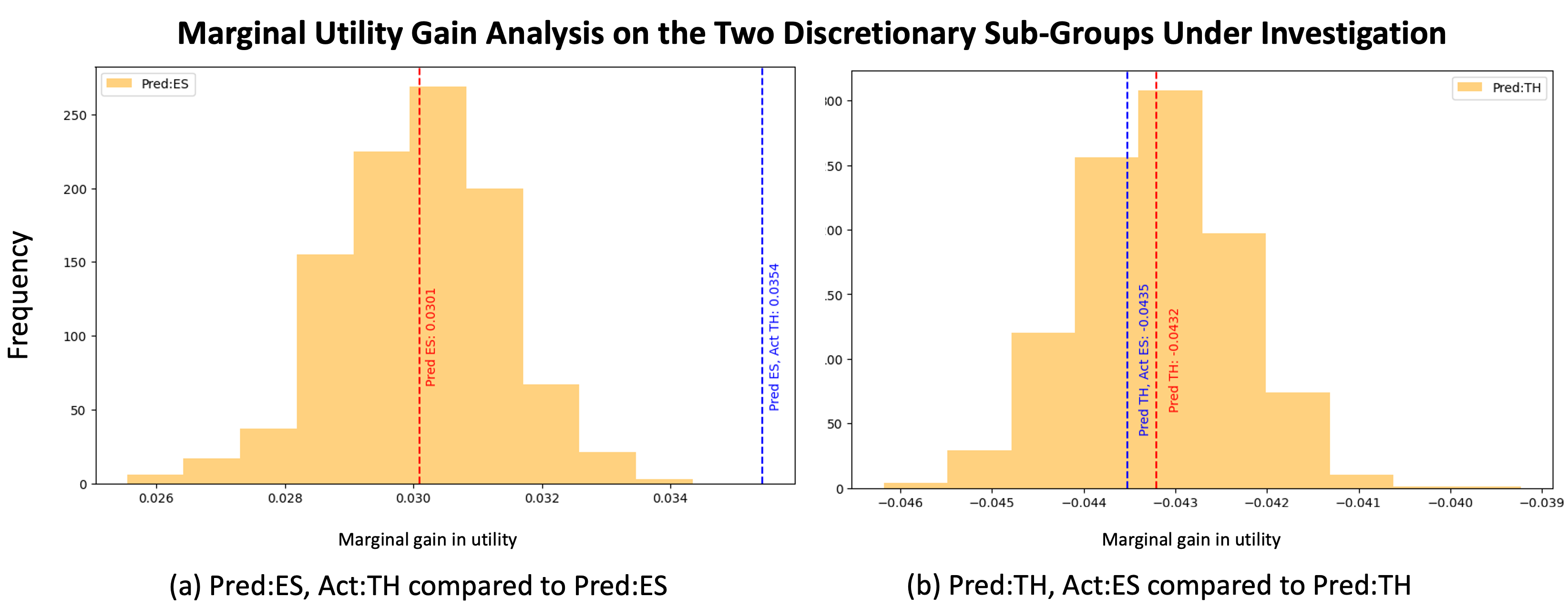}
    \caption{Non-parametric test comparing the marginal gain in utility resulting from the switch in interventions between the discretionary subgroups and the population predicted to receive the intervention.}
    \label{fig:marginal_gain_utility}
\end{figure}

\subsubsection{Marginal Benefit}


Turning our attention to Figure \ref{fig:marginal_gain_utility}, we are presented with analogous insights derived from the non-parametric tests applied to gauge marginal benefits. A noteworthy distinction surfaces when contrasting the two subgroup analyses. For the EStoTH discretionary group, there is a significant benefit (again, outside the distribution of random samples from those predicted to receive ES), while there is no corresponding significant loss for the THtoES discretionary subgroup. Specifically, the mean value within this subgroup of households nestles comfortably within the 34\textsuperscript{th} percentile spectrum when juxtaposed with the averages computed from the control group. This intriguing discovery suggests that caseworkers, even in scenarios where the imperative is to assign households a less intensive intervention, accomplish this task in a manner that circumvents substantial perturbations and maintains a certain level of stability. Impressively, this is achieved while also effectively harnessing discretionary actions to secure tangible gains, steering the transition towards a \emph{more} intensive intervention.


Synthesizing our discoveries within this segment, a coherent narrative emerges. Collectively, our findings shed light on the overarching trend: caseworker discretion, by and large, gravitates towards individuals of comparatively lower vulnerability. Delving deeper, within this subgroup, discretionary measures exhibit a distinct outcome-centric orientation. Caseworkers adeptly wield discretion to enhance outcomes, particularly in the context of the EStoTH transition, all while delicately managing the potential repercussions for the THtoES counterpart—those conceivably requiring reassignment to respect capacity constraints. Also, note that caseworkers themselves do not have access to any kind of outcome predictions, but still proficiently align their actions with external quantitative machine learning metrics. This alignment, predicated solely on their cumulative experience and judgment, underscores the congruence between human expertise and algorithmic evaluation.

\section{Discussion}


The integration of AI tools into various public administration domains, such as child welfare protection, social service delivery, and policing, has given rise to a critical inquiry into the roles that street-level bureaucrats play, both historically and prospectively, in shaping decision-making processes. Amidst the fervor of technological advancements, it is important to understand the intricate dynamics at play between automated systems and the human element inherent to public service interactions. Beyond the mere mechanization of tasks, a pivotal consideration is the form of judgment and discretion exercised by human agents, the  ``bureaucratic counterfactual'' \cite{johnson2022bureaucratic}. 
In this paper, we show that homelessness service providers were very consistent in their decision-making, alleviating possible concerns about procedural justice. At the same time, their discretionary behavior displays interesting nuances. They typically only target less vulnerable households to make discretionary decisions (choosing the ``standard'' decision for more vulnerable households). Still, within that, their decisions appear sophisticated and beneficial in maximizing marginal benefit. These results demonstrate consistency and positive applications of judgment. They can also inform how best to target areas of AI tool development in this domain to best assist these human experts in the future.

\section{Acknowledgments}

This work was partially supported by the NSF (IIS- 1939677, IIS-2127752) and by Amazon through an NSF FAI award. We thank local community partners in helping to conceptualize the challenges facing homeless service delivery.

\bibliography{main}

\newpage
\onecolumn 

\section{Supplementary Materials}
\subsection{Appendix A: Data Summary}

The dataset includes records on 12,715 households. The target variable here is a one-vs-all prediction label for each intervention, or in general terms what intervention a household receives. We do not include permanent supportive housing in our analysis because it was not assigned as a first-time intervention. A break-down of the number of intervention assignments in the data we analyze is shown below:

\begin{table}[H]
\centering
\begin{tabular}{|c|c|}
\hline
\textbf{Service Type} & \textbf{Number Assigned} \\ \hline
Emergency Shelter     & 4,441                    \\
Transitional Housing  & 2,451                    \\
Rapid Re-Housing      & 846                      \\
Prevention            & 4,977                    \\ \hline
\textbf{Total}        & \textbf{12,715}          \\ \hline
\end{tabular}%
\end{table}

\subsection{Appendix B: The features in the data}

A comprehensive description of all the features that we used in the creation of the decision trees is shown in the table below. All categorical variables were one-hot-encoded.

\begin{table}[H]
\resizebox{\textwidth}{!}{%
\begin{tabular}{|ll|}
\hline
\multicolumn{2}{|c|}{\textbf{Binary Features}}                                                                                                                                                      \\ \hline
\multicolumn{1}{|c|}{\textbf{Feature Name}}                  & \multicolumn{1}{c|}{\textbf{Description}}                                                                                            \\ \hline
\multicolumn{1}{|l|}{HUDChronicHomeless}                     & Whether or not he client is chronically homeless.                                                                                    \\
\multicolumn{1}{|l|}{Gender}                                 & In original documentation can take on one of  8 values. But in data is only either Male or Female.                                   \\
\multicolumn{1}{|l|}{SpousePresent}                          & Varaible indicating whether or not client has a spouse.                                                                              \\ \hline
\multicolumn{2}{|c|}{\textbf{Categorical Features}}                                                                                                                                                 \\ \hline
\multicolumn{1}{|c|}{\textbf{Feature Name}}                  & \multicolumn{1}{c|}{\textbf{Description}}                                                                                            \\ \hline
\multicolumn{1}{|l|}{PrimaryRace}                            & The primary race of client. Can be one of 7 categories.                                                                              \\
\multicolumn{1}{|l|}{Ethnicity}                              & The ethnicity of client. Can be one of 5 categories.                                                                                 \\
\multicolumn{1}{|l|}{PriorResidence}                         & The type of residence client had before entering system. Can be one of 25 categories.                                                \\
\multicolumn{1}{|l|}{VeteranStatus}                          & Varaible indicating whether or not client is a veteran. Can take one of 5 values.                                                    \\
\multicolumn{1}{|l|}{DisablingCondition}                     & Variable indicating whether or not client has a disabling condition. Can be one of 3 categories.                                     \\
\multicolumn{1}{|l|}{ReceivePhysicalDisabilityServices}      & Varaible indicating whether or not client received disability services. Can be one of 5 categories.                                  \\
\multicolumn{1}{|l|}{HasDevelopmentalDisability}             & Variable indicating whether or not client has developmental disability. Can be one of 5 categories.                                  \\
\multicolumn{1}{|l|}{ReceiveDevelopmentalDisabilityServices} & Varaible indicating whether or not client received services for developmental disability. Can be one of 5 categories.                \\
\multicolumn{1}{|l|}{HasChronicHealthCondition}              & Variable indicating whether or not client has a chronic health condition. Can be one of 5 categories.                                \\
\multicolumn{1}{|l|}{ReceiveChronicHealthServices}           & Varaible indicating whether or not client received services for a chronic health condition. Can be one of 5 categories.              \\
\multicolumn{1}{|l|}{HasHIVAIDS}                             & Variable indicating whether or not client has HIV\textbackslash{}AIDS. Can be one of 5 categories.                   \\
\multicolumn{1}{|l|}{ReceiveHIVAIDSServices}                 & Varaible indicating whether or not client received services for HIV\textbackslash{}AIDS. Can be one of 5 categories. \\
\multicolumn{1}{|l|}{HasMentalHealthProblem}                 & Variable indicating whether or not client has a mental health condition. Can be one of 5 categories.                                 \\
\multicolumn{1}{|l|}{ReceiveMentalHealthServices}            & Varaible indicating whether or not client received services related to mental health. Can be one of 5 categories.                    \\
\multicolumn{1}{|l|}{HasSubstanceAbuseProblem}               & Variable indicating whether or not client has a a substance abuse problem. Can be one of 5 categories.                               \\
\multicolumn{1}{|l|}{ReceiveSubstanceAbuseServices}          & Varaible indicating whether or not client received services related to substance abuse. Can be one of 5 categories.                  \\
\multicolumn{1}{|l|}{DomesticViolenceSurvivor}               & Variable indicating whether or not client is a survivor of domestic violence. Can be one of 5 categories.                            \\ \hline
\multicolumn{2}{|c|}{\textbf{Continuous Features}}                                                                                                                                                  \\ \hline
\multicolumn{1}{|c|}{\textbf{Feature Name}}                  & \multicolumn{1}{c|}{\textbf{Description}}                                                                                            \\ \hline
\multicolumn{1}{|l|}{Age}                                    & The age of the client                                                                                                                \\
\multicolumn{1}{|l|}{Calls}                                  & The number of calls made to hotline before entry                                                                                     \\
\multicolumn{1}{|l|}{Wait}                                   & The time (in days) elapsed since first call to hotline to intervention assignment                                                    \\
\multicolumn{1}{|l|}{RatioOfNumCallstoWaitTime}              & Ratio of number of calls to wait time. Calculated variable to measure `persistence'                                                  \\
\multicolumn{1}{|l|}{MonthlyAmount}                          & The total income (in dollars) of client per month.                                                                                   \\
\multicolumn{1}{|l|}{numMembers}                             & The number of members in the client's households.                                                                                    \\
\multicolumn{1}{|l|}{Children}                               & The number of children in client's household.                                                                                        \\
\multicolumn{1}{|l|}{Children 0-2}                           & The number of children aged 0 to 2 in client's household.                                                                            \\
\multicolumn{1}{|l|}{Children 3-5}                           & The number of children aged 3 to 5 in client's household.                                                                            \\
\multicolumn{1}{|l|}{Children 6-10}                          & The number of children aged 6 to 10 in client's household.                                                                           \\
\multicolumn{1}{|l|}{Children 11-14}                         & The number of children aged 11 to 14 in client's household.                                                                          \\
\multicolumn{1}{|l|}{Children 15-17}                         & The number of children aged 15 to 17 in client's household.                                                                          \\
\multicolumn{1}{|l|}{UnrelatedChildren}                      & The number of clidren in client's household not related to them.                                                                     \\
\multicolumn{1}{|l|}{UnrelatedAdults}                        & The number of adults in client's household not realted to them.                                                                      \\ \hline
\end{tabular}%
}
\end{table}
\newpage
\subsection{Appendix C: Creating Vulnerability Scores}
The vulnerability scores were calculated using the following table. It was created to replicate the VI-SPDAT as closely as possible given the features that we had in our data. 

\begin{table}[H]
\resizebox{\textwidth}{!}{%
\begin{tabular}{|ll|}
\hline
\multicolumn{1}{|c|}{\textbf{CRITERIAS}}                                                                                                                & \multicolumn{1}{c|}{\textbf{SCORES}} \\ \hline
\multicolumn{2}{|c|}{\textbf{Single Person Households}}                                                                                                                                        \\ \hline
\multicolumn{1}{|l|}{If single parent   with 2 or more children AND/OR a child aged 11 and younger}                                                     & +1                                   \\ \hline
\multicolumn{1}{|l|}{If NOT single   parent with 3 or more children AND/OR a child aged 6 and younger}                                                  & +1                                   \\ \hline
\multicolumn{2}{|c|}{\textbf{History of Housing and Homelessness}}                                                                                                                             \\ \hline
\multicolumn{1}{|l|}{If prior residence   is anything other than `emergency shelter' or `transitional housing for   homeless persons', or `safe haven'} & +1                                   \\ \hline
\multicolumn{1}{|l|}{If chronically   homeless}                                                                                                         & +1                                   \\ \hline
\multicolumn{2}{|c|}{\textbf{Risks}}                                                                                                                                                           \\ \hline
\multicolumn{1}{|l|}{Used a crisis   service i.e., if any one of the following is true:}                                                                &                                      \\
\multicolumn{1}{|l|}{- Received physical   disability services}                                                                                           &                                      \\
\multicolumn{1}{|l|}{- Received   developmental disability services}                                                                                      &                                      \\
\multicolumn{1}{|l|}{- Received chronic   health services}                                                                                                &                                      \\
\multicolumn{1}{|l|}{- Received HIV/AIDS   services}                                                                                                      & +1                                   \\
\multicolumn{1}{|l|}{- Received mental   health services}                                                                                                 &                                      \\
\multicolumn{1}{|l|}{- Received substance   abuse services}                                                                                               &                                      \\ \hline
\multicolumn{1}{|l|}{If prior residence   is prison}                                                                                                    & +1                                   \\ \hline
\multicolumn{2}{|c|}{\textbf{Wellness}}                                                                                                                                                        \\ \hline
\multicolumn{1}{|l|}{Physical health of   client i.e., if any one of the following is true:}                                                            &                                      \\
\multicolumn{1}{|l|}{- Has a chronic   health condition}                                                                                                  &                                      \\
\multicolumn{1}{|l|}{- Has HIV/AIDS and   Receives HIV/AIDS services}                                                                                     &                                      \\
\multicolumn{1}{|l|}{- Has physical   disability}                                                                                                         & +1                                   \\
\multicolumn{1}{|l|}{- Has developmental   Disability}                                                                                                    &                                      \\
\multicolumn{1}{|l|}{- Has substance abuse   problem}                                                                                                     &                                      \\ \hline
\multicolumn{1}{|l|}{Mental health of   client i.e., if any one of the following is true:}                                                              &                                      \\
\multicolumn{1}{|l|}{- Has mental health   problem}                                                                                                       & +1                                   \\
\multicolumn{1}{|l|}{- Is domestic   violence survivor}                                                                                                   &                                      \\ \hline
\end{tabular}%
}
\end{table}
\subsection{Appendix D: Intervention Prediction Using CART}

As mentioned in the text, we constructed trees using the CART algorithm in a manner similar to XGBoost. When using CART we performed a grid search based on the following hyper-parameters: 
\begin{itemize}
    \item Measure of impurity: Gini, Entropy 
    \item Max depth : $d \in \left[ 1, 10\right]$
    \item Minimum samples per split : $s \in \left[ 2, 10\right]$
    \item Minimum samples in a leaf node: $n \in \left[ 1, 5\right]$
\end{itemize}

The combination of hyper-parameters that yielded the best results for CART is reported in the table below: 

\begin{table}[H]
\centering
\begin{tabular}{|l|c|c|c|c|}
\hline
\multicolumn{1}{|c|}{\textbf{Intervention Type}} & \textbf{Criterion} & \textbf{Max Depth} & \textbf{Min Samples in Leaf} & \textbf{Min Samples in Split} \\ \hline
Emergency Shelter                                & Gini               & 9                  & 4                            & 8                             \\
Transitional Housing                             & Gini               & 7                  & 4                            & 8                             \\
Rapid Re-Housing                                 & Gini               & 4                  & 3                            & 2                             \\
Prevention                                       & Gini               & 8                  & 1                            & 9                             \\ \hline
\end{tabular}
\label{tab:experimental_details}
\end{table}

The ROC curves for the trees generated using CART is shown in the figure below: 

\begin{figure}[H]
    \centering
    \includegraphics[width=.5\columnwidth]{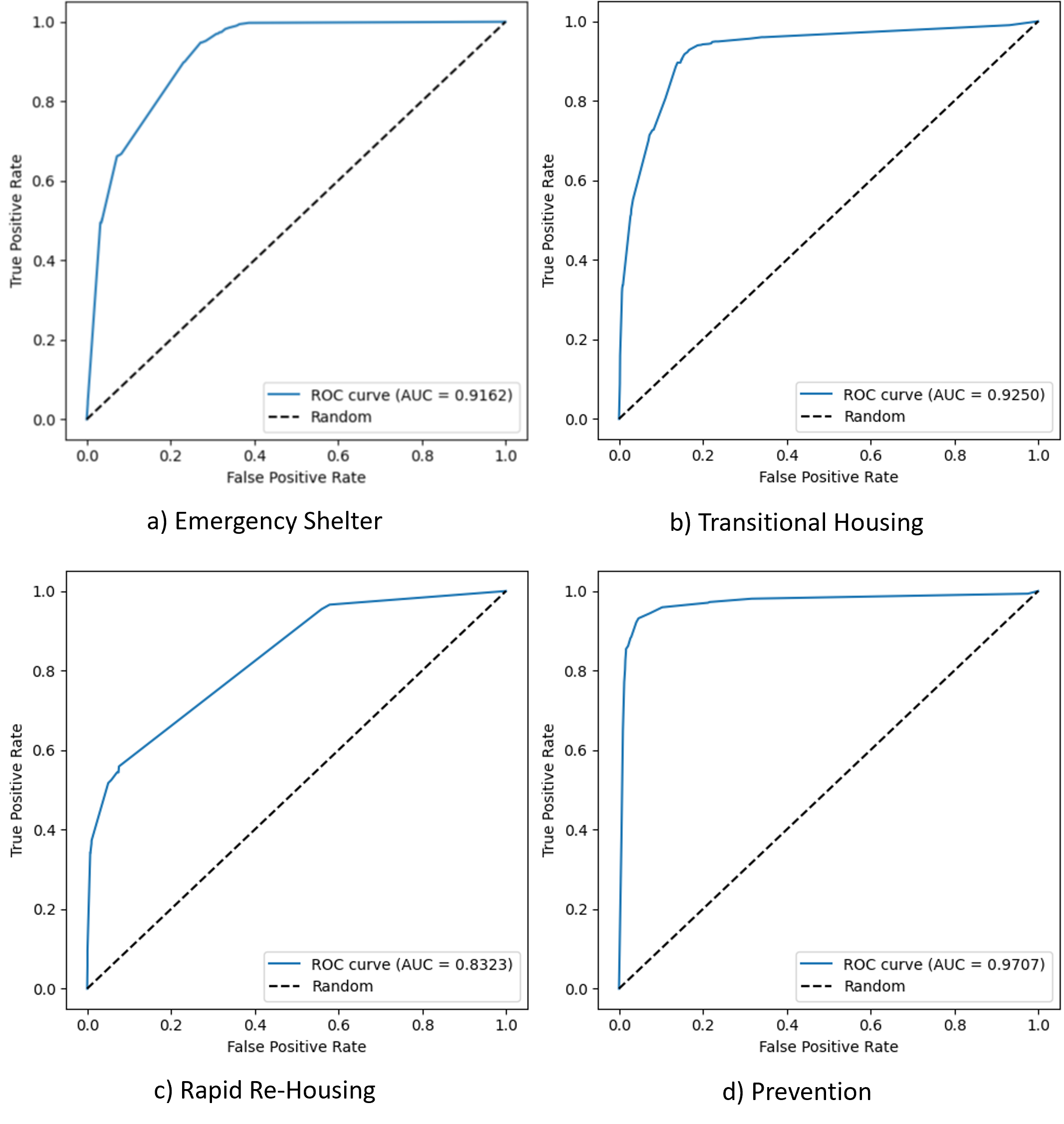}
    \caption{ROC curves for each tree used to learn intervention assignments using the CART algorithm. Predictive discrimination, measured using the area under the ROC curve is very high (similar to XGBoost), further implying significant consistency in caseworker decision-making.}
    \label{fig:cart_roc} 
\end{figure}

\noindent NOTE: Because we use a one-versus-all model for each intervention, it is possible that a single household might be predicted to receive more than one intervention. In the event that this happens, the most `confident' prediction is taken as the final predicted intervention for the household.  

\end{document}